\documentclass[runningheads]{llncs}

\usepackage{eccv}

\usepackage{eccvabbrv}

\usepackage{graphicx}
\usepackage{tabularx}
\usepackage{multicol}
\usepackage{multirow}
\usepackage{booktabs}
\usepackage{algorithm}
\usepackage{algpseudocode}
\usepackage{amsmath}

\usepackage[accsupp]{axessibility}  

\usepackage{hyperref}

\usepackage{orcidlink}

\begin{document}

\title{Meta Learning-Driven Iterative Refinement for Robust Anomaly Detection in Industrial Inspection} 

\titlerunning{MLD-IR for Robust Anomaly Detection in Industrial Inspection}

\author{Muhammad Aqeel\inst{1, 2}\orcidlink{0009-0000-5095-605X} \and
Shakiba Sharifi\inst{1} \and
Marco Cristani\inst{1}\orcidlink{0000-0002-0523-6042} \and
Francesco Setti\inst{1}\orcidlink{0000-0002-0015-5534}}

\authorrunning{M.~Aqeel et al.}

\institute{Dept. of Engineering for Innovation Medicine, University of Verona\\ 
Strada le Grazie 15, Verona, Italy \and
\email{muhammad.aqeel@univr.it}\\
\url{https://www.dimi.univr.it/}
}

\maketitle

\begin{abstract}
  This study investigates the performance of robust anomaly detection models in industrial inspection, focusing particularly on their ability to handle noisy data. We propose to leverage the adaptation ability of meta-learning approaches to identify and reject noisy training data to improve the learning process. In our model, we employ Model Agnostic Meta-Learning (MAML) and an iterative refinement process through an Inter-Quartile Range rejection scheme to enhance their adaptability and robustness. This approach significantly improves the models' capability to distinguish between normal and defective conditions. Our results of experiments conducted on well-known MVTec and KSDD2 datasets demonstrate that the proposed method not only excels in environments with substantial noise but can also contribute in case of a clear training set, isolating those samples that are relatively out of distribution, thus offering significant improvements over traditional models.
\end{abstract}

\section{Introduction}
\label{sec:intro}

Anomaly detection is pivotal in industrial settings, particularly for ensuring rigorous quality control and effective defect detection in manufacturing processes. Traditional methods, while foundational, often struggle with the localization and precise identification of anomalies within images, which is paramount for modern industrial contexts such as automated inspection systems. 

Recent advancements in anomaly detection have increasingly utilized deep learning techniques that offer superior performance over traditional image-based methods. These new approaches are capable of detecting, localizing, and classifying anomalies with high precision, even at the pixel level. This shift is largely due to the development of convolutional neural networks (CNNs) and other deep learning frameworks that have been specifically adapted for anomaly detection in image data, demonstrating remarkable success in identifying subtle and complex anomalies that were previously challenging to detect with conventional methods \cite{bergmann2020uninformed, chow2020anomaly}.


The rarity and diversity of anomalies pose significant challenges as they require detection systems to operate with minimal human-labeled data. This scarcity makes the collection and annotation of a comprehensive dataset both expensive and time-consuming. To address these challenges, recent methods have employed unsupervised and semi-supervised learning techniques, where the models are trained predominantly on normal data and anomalies are identified as deviations from this learned norm \cite{chandola2009anomaly}.

Techniques such as autoencoders and generative adversarial networks (GANs) have been pivotal in this respect. Autoencoders, for instance, learn to reconstruct normal images and can highlight anomalies through reconstruction errors. Similarly, GANs have been adapted to generate normal data distributions, with anomalies identified by their failure to conform to these distributions \cite{schlegl2017unsupervised, akcay2019ganomaly}.

Furthermore, the integration of transfer learning and meta-learning into anomaly detection systems has shown promising results in reducing the need for extensive labeled datasets. These methods enable systems to adapt quickly to new types of anomalies or changes in data distribution, enhancing their flexibility and utility in dynamic industrial environments \cite{hu2020efficient, kang2021person}.

\begin{figure}[t]
\centering
\subfloat[Clean training]{\includegraphics[width=0.3\textwidth]{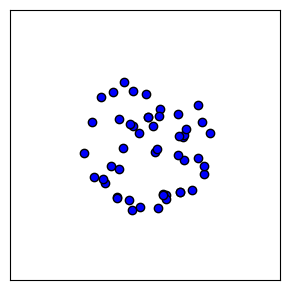}\label{fig:clean}}
\hfill
\subfloat[Corrupted training]{\includegraphics[width=0.3\textwidth]{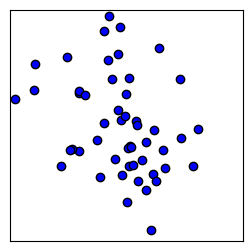}\label{fig:corrupt}}
\hfill
\subfloat[Iteratively refined training]{\includegraphics[width=0.3\textwidth]{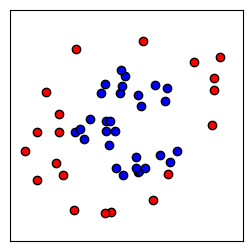}\label{fig:ourtraining}}
\hfill
\subfloat[Model learned with clean training data]{\includegraphics[width=0.3\textwidth]{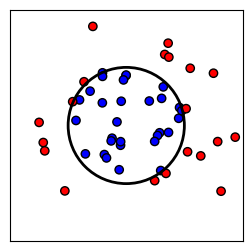}\label{fig:test}}
\hfill
\subfloat[Model learned with corrupted training data]{\includegraphics[width=0.3\textwidth]{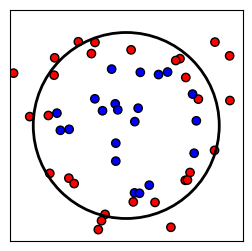}\label{fig:test1}}
\hfill
\subfloat[Model learned with iteratively refined data]{\includegraphics[width=0.3\textwidth]{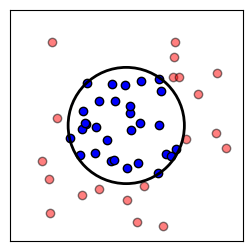}\label{fig:ourtest}}
\caption{Comparison of anomaly detection models under various training conditions: Clean, Corrupted, and Iteratively Refined. This sequence demonstrates the effectiveness of our metalearning-driven iterative refinement process in improving the robustness and accuracy of anomaly detection in industrial applications.}
\label{fig:ourmethod}
\end{figure}

In anomaly detection, the quality of training images plays a pivotal role in the effectiveness of the resulting model. In Figure \ref{fig:ourmethod} visually demonstrates how data quality impacts model performance, comparing clean, corrupted, and iteratively refined training scenarios. Figures \ref{fig:clean} and \ref{fig:corrupt} contrast clean and corrupted training scenarios, highlighting the challenges introduced by noise and anomalies in training data. Subsequently, Figure \ref{fig:ourtraining} introduces our method of iteratively refined training, designed to enhance data quality and robustness against such corruptions. 
The models trained on clean and corrupted data exhibit varying levels of accuracy and sensitivity, as shown in Figures \ref{fig:test} and \ref{fig:test1}, respectively. Our model, demonstrated in Figure \ref{fig:ourtest}, is tested under conditions reflecting the iterative training process, showcasing its superior capability to maintain high performance even in the presence of data imperfections. These visual comparisons underline the necessity and effectiveness of our proposed meta-learning driven iterative refinement approach in handling real-world industrial data complexities.

Addressing these challenges, our research introduces a robust, unsupervised anomaly detection framework that employs meta-learning with iterative refinement, uniquely tailored to address the complexity and variability inherent in industrial data settings, enhancing accuracy and reducing false alarms. This novel approach significantly improves the model's ability to adapt and recognize subtle, evolving anomaly patterns, crucial for detecting rare and nuanced anomalies. Our main contributions include:
\begin{itemize}
    \item We introduce a meta-learning framework that dynamically adapts to evolving industrial data characteristics, effectively overcoming the limitations of traditional statistical and machine learning methods.
    \item Through iterative refinement, the model continuously updates and fine-tunes its detection algorithms, leveraging new anomaly data to refine its understanding and improve accuracy.
    \item We specifically enhance the framework's adaptability to environments characterized by significant background noise and operational variability. This targeted adaptation focuses on maintaining high detection performance in noisy conditions, showcasing the framework's practical effectiveness and reliability in challenging industrial settings.
    \item We report extensive results on public benchmarks to prove the effectiveness of our approach to isolate problematic samples at training time and train a model on the surviving ones.
\end{itemize}

This paper is structured as follows: a review of related work in industrial anomaly detection is presented in Sect. \ref{sec:related}, we then detail the proposed methodology in Sect. \ref{sec:method}, and we demonstrate the approach's effectiveness in Sect. \ref{sec:experiment} and Sect. \ref{sec:results}. Finally, in Sect. \ref{sec:conclusion} we summarize our findings and discuss some future research directions.

\section{Related Work}
\label{sec:related}
This section reviews recent advancements in image-based anomaly detection, highlighting the transition from traditional statistical methods to sophisticated deep learning models that handle complex datasets. Modern approaches, specifically tailored for handling high-dimensional and complex image datasets, have revolutionized anomaly detection capabilities in industrial settings. This section reviews these developments, placing a strong emphasis on methods that excel in identifying subtle and intricate anomalies in visual data.

Convolutional Neural Networks (CNNs) have become the cornerstone of modern image-based anomaly detection due to their ability to learn hierarchical features from raw images automatically. CNNs are particularly effective in industrial inspection tasks, such as detecting defects in manufacturing processes, demonstrating superior performance over traditional methods \cite{bergmann2019mvtec}. However, their effectiveness can be limited by the quality and volume of training data. Generative Adversarial Networks (GANs) further enhance these capabilities by generating realistic images to identify anomalies as deviations that the generator network fails to reproduce accurately \cite{schlegl2017unsupervised}.

The integration of unsupervised learning techniques, such as Autoencoders and Isolation Forest, has broadened the scope of anomaly detection. Autoencoders excel by reconstructing normal data patterns and flagging significant reconstruction errors as anomalies, making them effective in noisy environments \cite{sakurada2014anomaly}. Isolation Forest uses random forests to isolate outliers, offering robust solutions where anomalies are subtle \cite{liu2008isolation}.

The use of semi-supervised methods like Deep SVDD, which learns a hypersphere enclosing normal data points with anomalies lying outside this boundary, represents a significant advancement in utilizing limited labeled data to enhance detection performance \cite{ruff2018deep}. Similarly, self-supervised anomaly detection techniques that leverage anomalous training data via iterative latent token masking have shown promise in enhancing model adaptability and performance \cite{patel2023self}. Informative knowledge distillation techniques are also being explored to improve deep anomaly detection models by preserving essential characteristics during model training, thus addressing some of the limitations inherent in traditional and older machine learning methods \cite{cao2022informative}.

While newer machine learning methods have taken center stage, traditional techniques such as K-means clustering and Principal Component Analysis (PCA) continue to provide a foundational understanding, albeit with limitations in complex image-based scenarios. K-means and PCA often struggle with the high dimensionality typical of industrial image data, highlighting the need for more adaptable and robust solutions \cite{hodge2004survey, shyu2003novel}.

Building on the foundational work reviewed here, our research integrates robust capabilities of deep learning with meta-learning and iterative refinement techniques, tackling challenges of noise and variability in image data while adapting dynamically to new anomaly patterns. By leveraging the strengths of both traditional and modern techniques, our framework sets a new benchmark for reliability and efficiency in industrial anomaly detection, ensuring high adaptability and performance even in environments characterized by significant background noise and operational variability.

\section{Method}
\label{sec:method}

%


Our methodology leverages the strengths of meta-learning, specifically Model Agnostic Meta Learning (MAML)~\cite{finn2017model}, to enhance the robustness of unsupervised anomaly detection in environments laden with noisy data. Meta-learning is particularly suited for these environments as it prepares the model to efficiently handle diverse datasets and adapt dynamically to unseen data conditions without extensive retraining. MAML achieves this by initializing model parameters that are predisposed to adjust to high variability and noise, a common challenge in real-world datasets.

This approach is meticulously outlined in Algorithm~\ref{algo}, which details the steps for continuous adaptation and improvement. The algorithm, as outlined, systematically enhances detection capabilities by recalibrating model parameters in response to real-time data inputs and outcomes, thereby progressively enhancing the model’s sensitivity to subtle and complex anomalies. By leveraging meta-learning, we ensure that our system not only starts with a robust foundation but also continues to adapt and improve, maintaining high accuracy and reliability. Our experiments, which harness the challenging conditions of the KSDD2 and MVTec AD dataset, demonstrate substantial improvements in detection performance over traditional methods, confirming the effectiveness of our meta-learning-enhanced iterative refinement process in practical, noisy scenarios.

\begin{algorithm}[H]
\caption{Iterative Refinement for Anomaly Detection}
\begin{algorithmic}[1]
\State Initialize model parameters $\theta$ with MAML
\Repeat
    \State \textbf{for} each mini-batch $B$ \textbf{do}
    \State \quad Compute feature extraction $f_{\text{ex}}(x;\phi)$ for $x \in B$
    \State \quad Transform features with $f_{\text{ML}}$ to compute scores $S(x)$
    \State \quad Compute IQR and adjust threshold T
    \State \quad Classify data points in $B$ based on threshold T
    \State \quad Update model parameters $\theta$ using MAML
    \State \textbf{end for}
    \State Evaluate model on validation set
\Until{performance on validation set converges}
\end{algorithmic}\label{algo}
\end{algorithm}

\subsection{Model Agnostic Meta Learning}
We implement Model Agnostic Meta Learning (MAML) to optimize both the initialization and continuous adaptation of model parameters. This dual approach not only prepares the system for initial exposure to operational data but also facilitates ongoing learning and adaptation based on evolving data patterns and operational feedback. This method is crucial in environments characterized by significant noise levels, where traditional learning algorithms might fail. The inner loop updates the model parameters on a small subset of data, while the outer loop optimizes these updates across multiple subsets, ensuring generalized performance. We implement Model-Agnostic Meta-Learning (MAML) to address the challenge of rapid adaptation under such noisy conditions. The core principle of MAML is to discover an optimal set of initial model parameters, $\theta$, which can be efficiently fine-tuned through a minimal number of gradient updates to enhance performance on new tasks.

The implementation begins with the inner loop optimization, where the model parameters are adjusted specifically for each noisy dataset. This adjustment involves a gradient descent update defined as:

\begin{equation}
\theta' = \theta - \alpha \nabla_{\theta} L(S, \theta)
\end{equation}

where $\alpha$ is the learning rate, and $L(S, \theta)$ represents the loss computed on a support set $S$ that includes noise-distorted data. Here, $\theta'$ denotes the updated parameters after one gradient step, aimed at improving the model's performance specifically for the noise characteristics of $S$.

Following the inner loop, the adapted parameters $\theta'$ are evaluated on a corresponding query set $Q$ to calculate the task-specific loss. This step is crucial as it assesses the effectiveness of the inner loop adaptations against new samples of noisy data, providing a feedback mechanism to refine the model adaptation. The task-specific loss is referred as $
L(Q, \theta')$.

\begin{figure}[ht]
    \centering
    \includegraphics[width=1.0\linewidth, height=3cm]{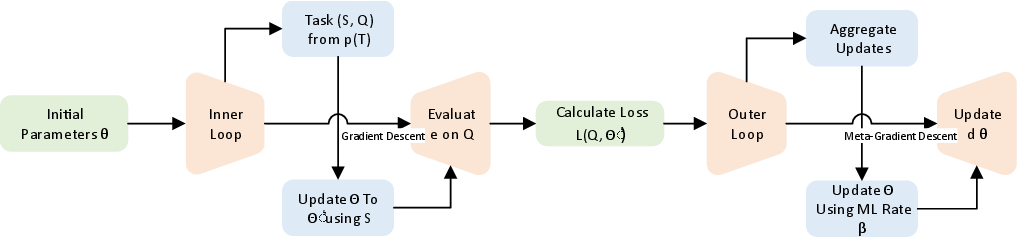}
    \caption{A schematic representation of the Model-Agnostic Meta-Learning (MAML) process, illustrating the inner and outer loop optimizations and highlighting adaptive adjustments to the model parameters $\theta$.}
    \label{fig:maml}
\end{figure}

\begin{figure}[H]
    \centering
    \includegraphics[trim={3cm 0 0 0},clip,width=\textwidth]{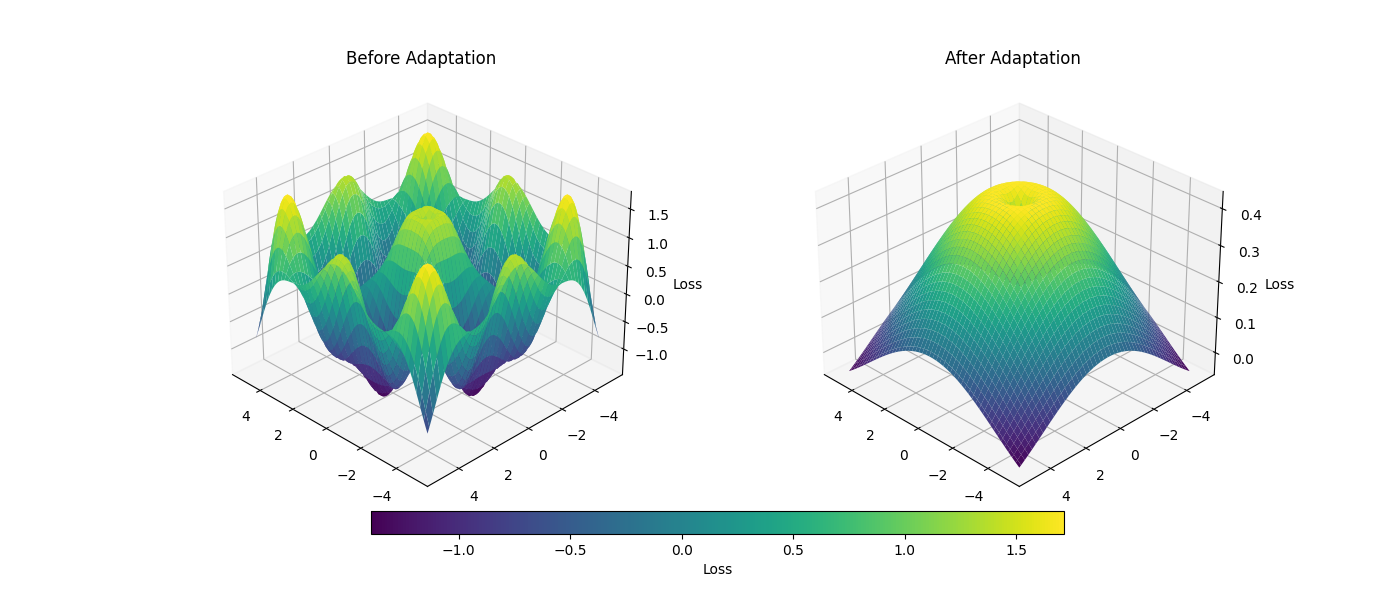}
    \caption{Comparison of loss landscapes before and after MAML adaptations, clearly showing how MAML smooths the optimization landscape, facilitating more effective learning and generalization.}
    \label{fig:loss}
\end{figure}

In the outer loop optimization, the initial model parameters $\theta$ are updated across multiple tasks. This stage aims to optimize a meta-objective that reduces the expected loss on new noisy datasets, thereby enhancing the model’s generalizability and robustness to variations in noise. The meta-objective is mathematically expressed as:

\begin{equation}
\theta = \theta - \beta \nabla_{\theta} \mathbb{E}_{(S, Q) \sim p(T)} L(Q, \theta')
\end{equation}
where $\beta$ is the meta-learning rate, and $p(T)$ denotes the distribution of tasks. This equation highlights the process of updating the initial parameters $\theta$ based on the aggregate feedback from multiple tasks, ultimately steering $\theta$ towards values that offer better general performance on noisy data. These theoretical enhancements are best visualized through direct comparison of optimization landscapes before and after its application, as illustrated in Figures~\ref{fig:maml} and \ref{fig:loss}.
To better visualize the process described, Figure~\ref{fig:maml} provides a schematic overview of the MAML framework, illustrating both the inner and outer optimization loops. This graphical representation underscores how MAML effectively tailors the learning process to maximize performance in dynamic, noisy environments.

Figure~\ref{fig:loss} displays the loss landscapes before and after the application of MAML adaptations, showing the complex, rugged landscape prior to adaptation and a significantly smoothed landscape post-adaptation. This visual comparison underscores the practical benefits of employing such sophisticated meta-learning techniques in high-noise environments.

Through this methodology, our implementation leverages MAML’s framework to minimize performance degradation in noisy conditions, ensuring rapid adaptation and high accuracy across various adversarial environments. This approach demonstrates how initial parameter settings can be fine-tuned to accommodate noise challenges, making the model more flexible and resilient.

\subsection{Iterative Refinement}
Our system adopts an integrated approach to anomaly detection, which not only leverages advanced scoring mechanisms and outlier removal techniques but also incorporates a meta-learning framework that enhances model adaptability and robustness. This iterative refinement process ensures precise anomaly identification and contributes significantly to the reliability and accuracy of our data analysis.

\subsubsection{Refinement Loop:} 
The iterative refinement process is outlined in Algorithm~\ref{algo}, which systematically recalibrates model parameters in response to real-time data inputs and outcomes. This process progressively enhances the model’s sensitivity to subtle and complex anomalies, ensuring that our system maintains high accuracy and reliability. Our experiments with the KSDD2 and MVTec AD datasets, known for their challenging conditions, demonstrate substantial improvements in detection performance over traditional methods. Inspired by the foundational model that uses a pre-trained feature extractor~\cite{rudolph2021same}, our approach extends its adaptability, applying learned insights across multiple data scenarios without additional optimization of the feature extractor.

\subsubsection{Score Computation:}
Initially, anomaly scores are computed using a hybrid metric, which integrates the feature distribution from the pretrained extractor and outputs from the meta-learning model. We define the score computation as follows:

\begin{equation}
S(x) = \mathbb{E}_{\phi}[-\log p_Z(f_{\text{ML}}(f_{\text{ex}}(x;\phi)))]
\end{equation}

Here, \( p_Z \) represents the probability density function in the latent space \( Z \), \( f_{\text{ex}} \) denotes the feature extraction from a pretrained model, \( f_{\text{ML}} \) is the transformation function from the meta-learning model, and \( \phi \) are the parameters optimized by MAML. This scoring method captures deviations from a predefined distribution, essential for the preliminary identification of potential outliers.

\subsubsection{Integration with Density Estimation:}
Our anomaly detection framework further integrates a scoring function based on density estimation from a normalizing flow transformation, aligning with traditional models. The scoring function for an image \( x \) subjected to transformations \( S_i \in S \) is defined as:
\begin{equation}
\tau(x) = \mathbb{E}_{S_i \in S}[-\log p_Z(f_{\text{NF}}(f_{\text{ex}}(S_i(x))))]
\end{equation}
This formulation ensures the model captures anomalies by evaluating the negative log-likelihoods across multiple transformations (e.g., rotations, brightness, and contrast adjustments), thus providing a robust mechanism to detect anomalies effectively.

\subsubsection{Dynamic Thresholding:}
After initial scoring, we use the Interquartile Range (IQR) for further data refinement. The IQR helps in filtering out noise and less relevant features through dynamic thresholding. In the thresholding process, we employ a scaling factor \( k \) to adjust the sensitivity of our anomaly detection according to the specific characteristics and variability of the data observed. The scaling factor \( k \) is adjusted to optimize the balance between false positives and false negatives, crucial for the effectiveness of anomaly detection systems. The upper threshold is formulated as follows:

\begin{equation}
T = Q_3 + k \times \text{IQR}
\end{equation}

where \( Q_3 \) is the third quartile and \( \text{IQR} = Q_3 - Q_1 \) is the interquartile range. This dynamically calculated threshold adapts to changes in anomaly score distributions, enhancing the model’s capability to differentiate effectively between normal and anomalous data.

\section{Experiments}
\label{sec:experiment}
This section outlines the experimental results achieved from applying our methodology to anomaly detection tasks in the presence of noisy training data. The experiments aimed to assess the enhancement in detection precision and stability across different industrial scenarios.

We began our experiments with the MVTec-AD dataset~\cite{bergmann2019mvtec}, a well-established benchmark featuring 15 distinct industrial product types. This dataset is particularly notable for its diversity, including more than 70 different defect types each meticulously identified with specific labels and segmentation masks.  

Subsequently, we utilized the KSDD2 dataset~\cite{Bozic2021COMIND}, sourced from the Kolektor Group d.o.o. This dataset includes RGB images of defective production items, characterized by a broad range of defects varying in size, shape, and color—from minor scratches to severe surface damage. 

For both datasets, we conducted experiments using proportionally sampled training and testing sets. The testing set consisted of a balanced distribution of nominal and anomalous samples. Each test was repeated multiple times to ensure statistical reliability, and average results were reported. To simulate noisy conditions inherent in unsupervised learning environments, the proportion of challenging data within the training set was incrementally adjusted. This adjustment was carried out at noise levels from 0\% to 50\%, progressively increasing the complexity of the training data to test the robustness of our detection algorithm.



We implemented our code using PyTorch framework. We resized images to \(448 \times 448\) pixels and applied standardized preprocessing techniques, including optional rotations and normalization. 
The network architecture incorporates three scales of input resolution, includes 8 coupling blocks in the normalizing flow model, and features scale-translation networks with fully connected layers, each comprising 2048 neurons, without dropout. We set the training parameters with a learning rate of \(2 \times 10^{-4}\) and conducted training over 240 epochs with a batch size of 96. 


We compared our proposed methodology with existing models including DifferNet~\cite{rudolph2021same}, One Shot Removal (OSR)~\cite{10.1007/978-3-031-66705-3_11}, and Iterative Refinement Process (IRP)~\cite{aqeel2024self}. DifferNet leverages a multi-scale feature extraction process to assign meaningful likelihoods to images, thereby facilitating defect localization. DifferNet represents also the model for feature extraction used in our procedure. OSR enhances the robustness of surface defect detection models through a novel training pipeline, which involves initial training, anomalous sample removal, and model fine-tuning. The IRP strategy systematically refines the training dataset through iterative cycles, enhancing detection accuracy and model robustness, especially in noisy environments.

We tested our model's capability to handle varying degrees of complexity in the training data, ranging from 0\% to 50\% noise levels. These levels represent the proportion of challenging samples included to simulate noisy conditions typically encountered in unsupervised learning scenarios. The model’s robustness was evaluated by measuring its performance using the AUROC score across these noise levels. The AUROC score provides a reliable indicator of the model's ability to differentiate between nominal and anomalous behaviors under increasing noise conditions. Each experiment was repeated three times to confirm the results' consistency and to comprehensively assess the impact of noise on detection accuracy.

\section{Results}
\label{sec:results}

This section presents the detailed evaluation results of four different anomaly detection models applied to the MVTec AD dataset across various classes, followed by an overall performance analysis that includes results from both the MVTec AD and KSDD2 datasets. The effectiveness of each model is assessed primarily using the AUROC metric, which measures their capability to differentiate between normal and defective samples under varying levels of noise.

\subsection{Performance in MVTec AD Classes}
\begin{figure}[tp]
\centering
\subfloat[Bottle]{\includegraphics[width=0.49\textwidth, height=3cm]{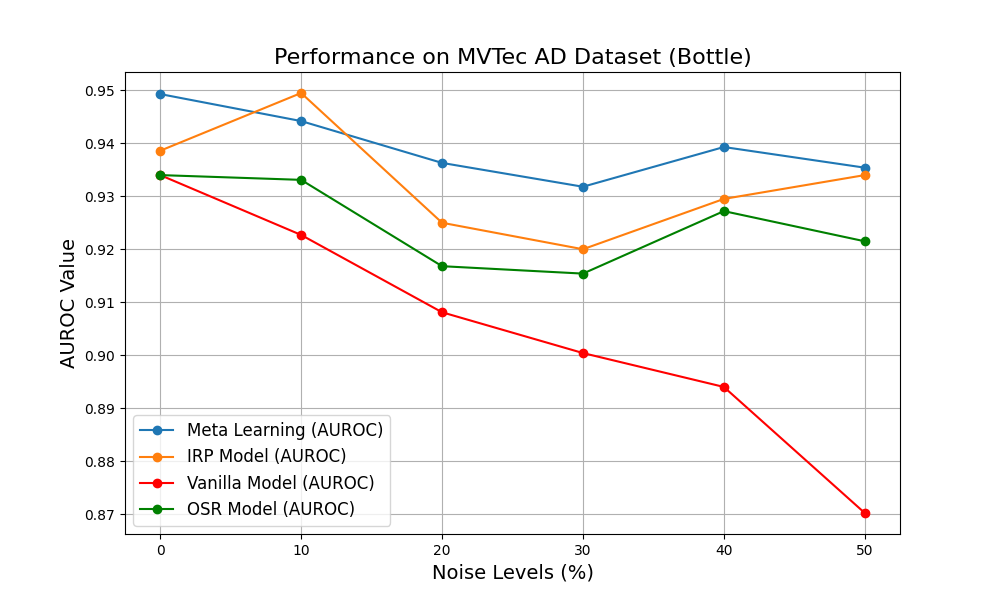}\label{fig:bottle}}
\hfill
\subfloat[Cable]{\includegraphics[width=0.49\textwidth, height=3cm]{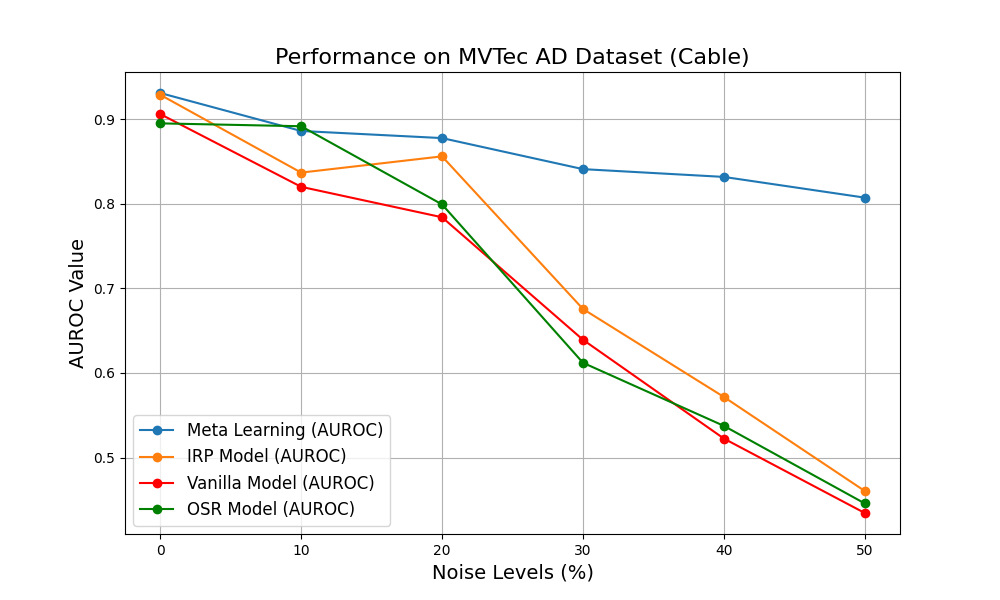}\label{fig:cable}}
\hfill
\subfloat[Capsule]{\includegraphics[width=0.49\textwidth, height=3cm]{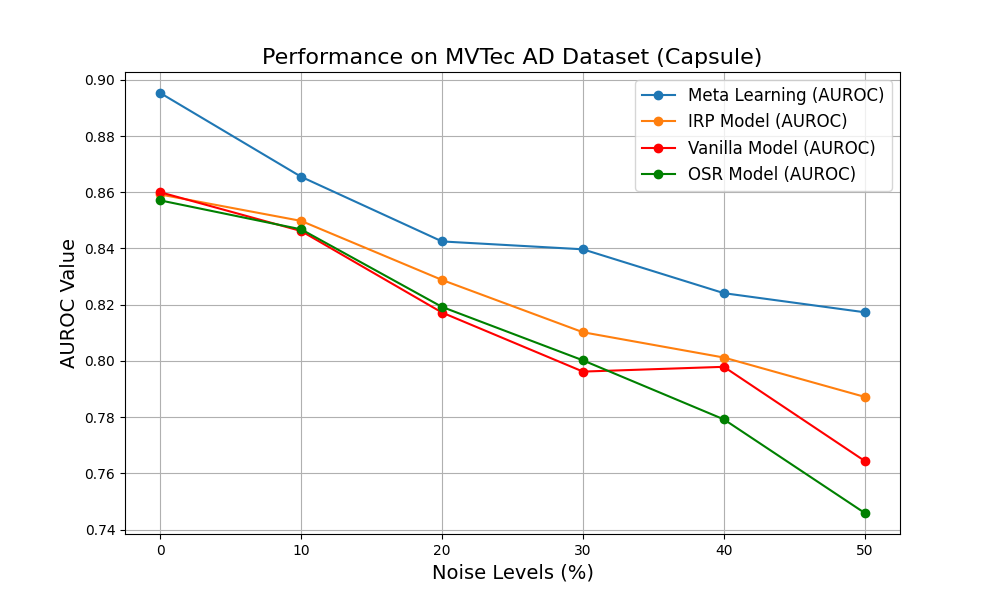}\label{fig:capsule}}
\hfill
\subfloat[Carpet]{\includegraphics[width=0.49\textwidth, height=3cm]{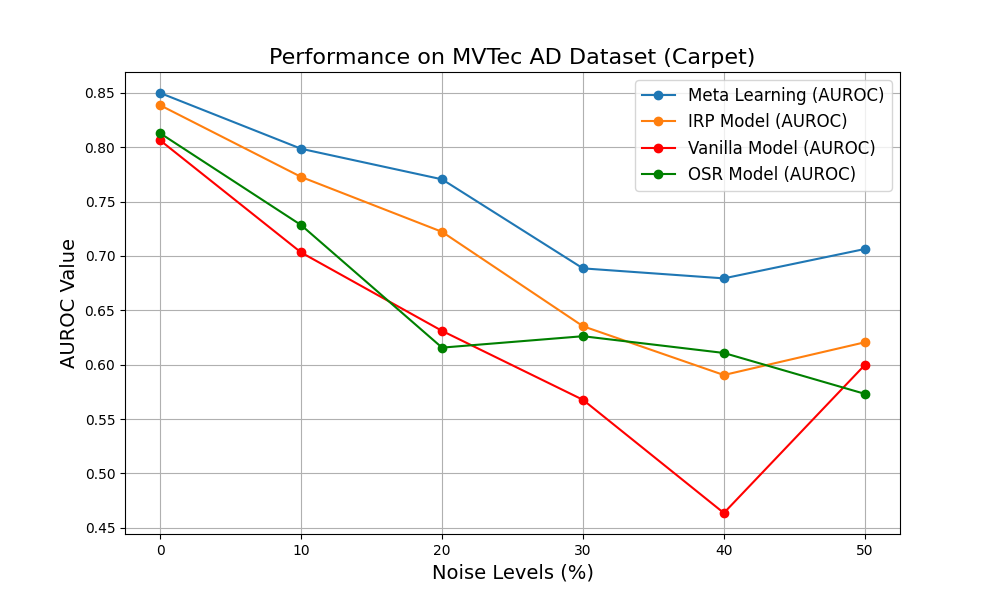}\label{fig:carpet}}
\hfill
\subfloat[Leather]{\includegraphics[width=0.49\textwidth, height=3cm]{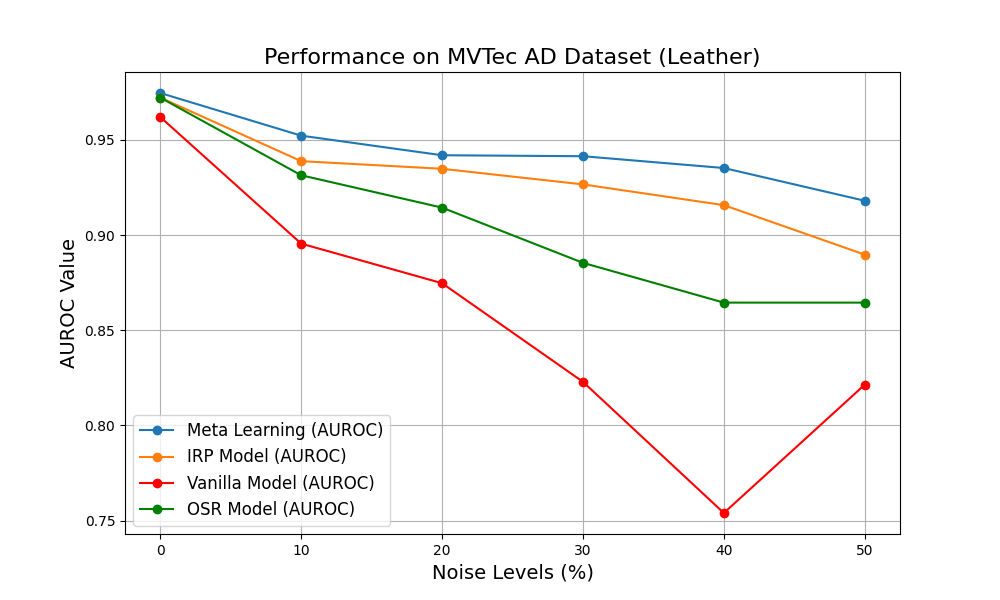}\label{fig:leather}}
\hfill
\subfloat[MetalNut]{\includegraphics[width=0.49\textwidth, height=3cm]{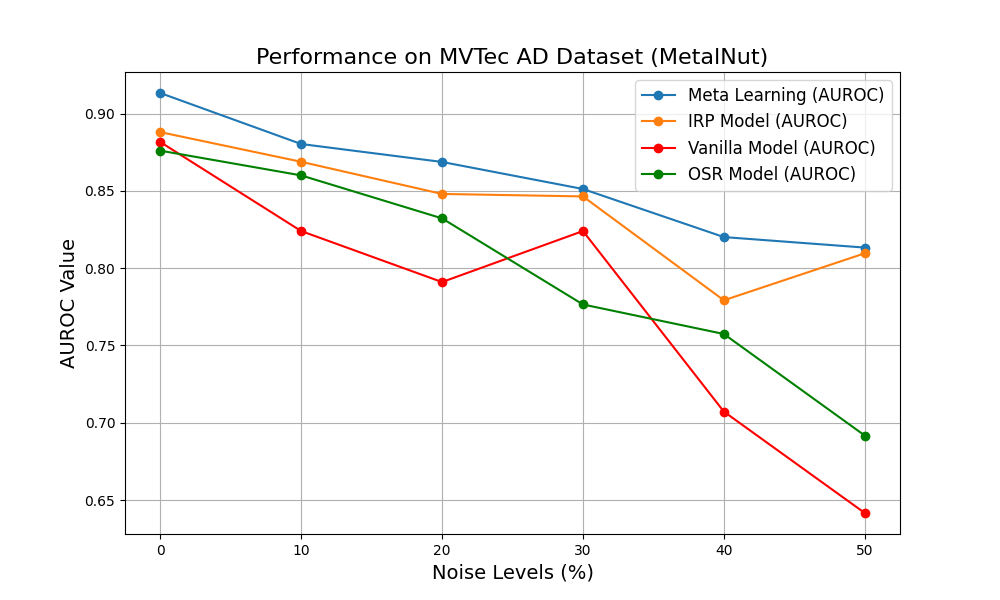}\label{fig:metalnut}}
\hfill
\subfloat[Pill]{\includegraphics[width=0.49\textwidth, height=3cm]{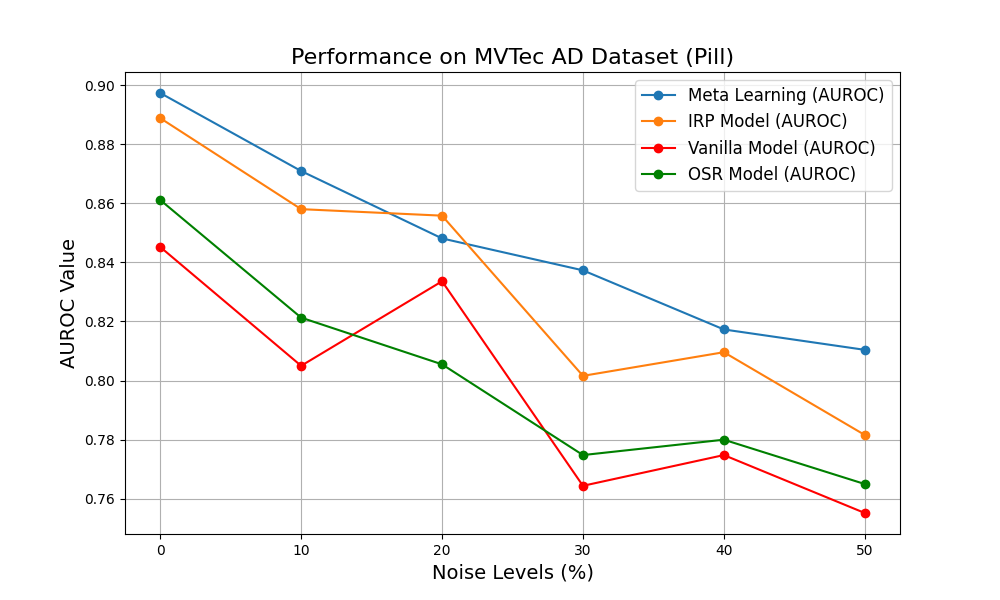}\label{fig:pill}}
\hfill
\subfloat[Screw]{\includegraphics[width=0.49\textwidth, height=3cm]{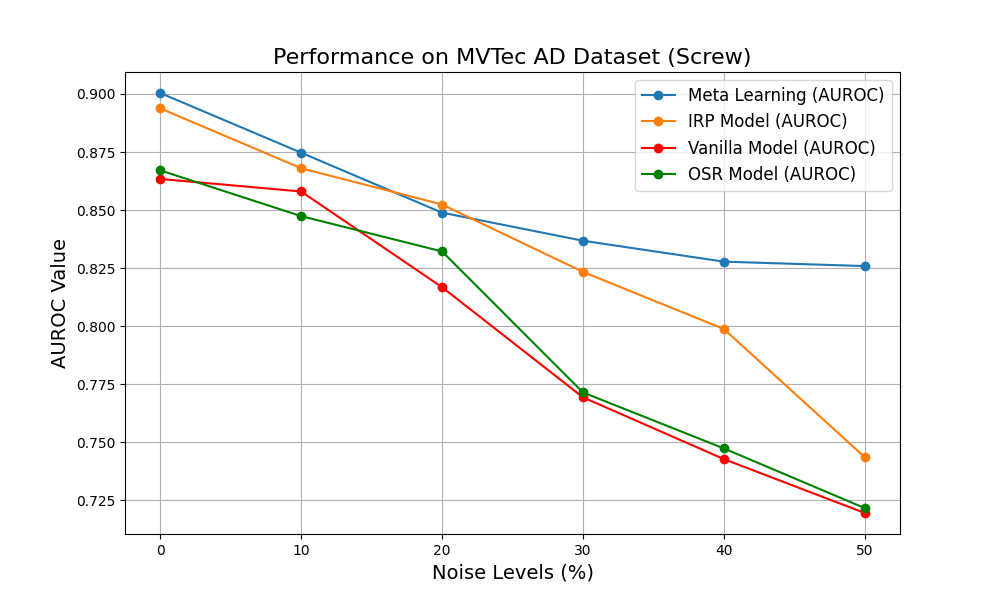}\label{fig:screw}}
\hfill
\subfloat[Tile]{\includegraphics[width=0.49\textwidth, height=3cm]{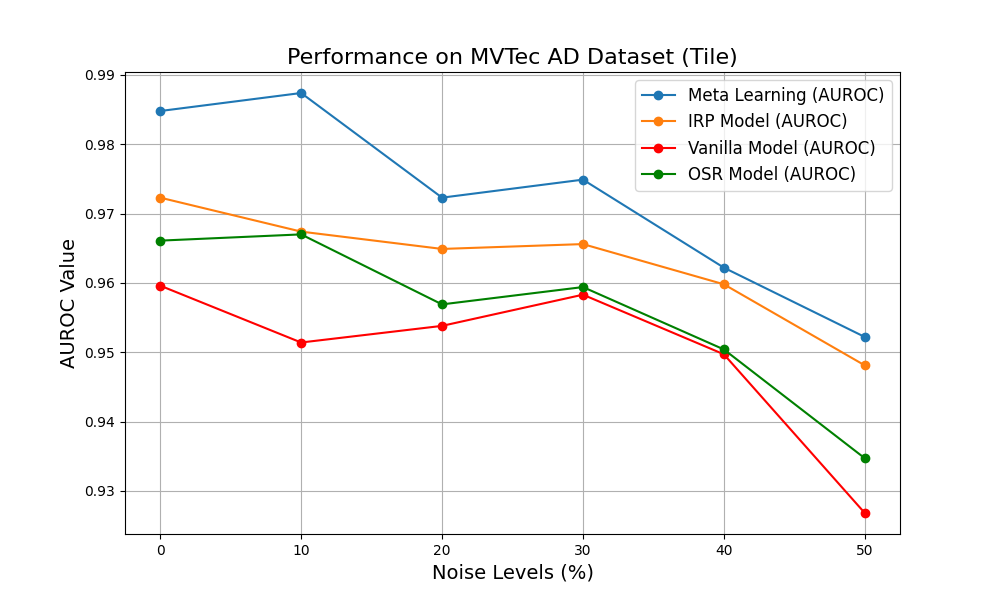}\label{fig:tile}}
\hfill
\subfloat[Zipper]{\includegraphics[width=0.49\textwidth, height=3cm]{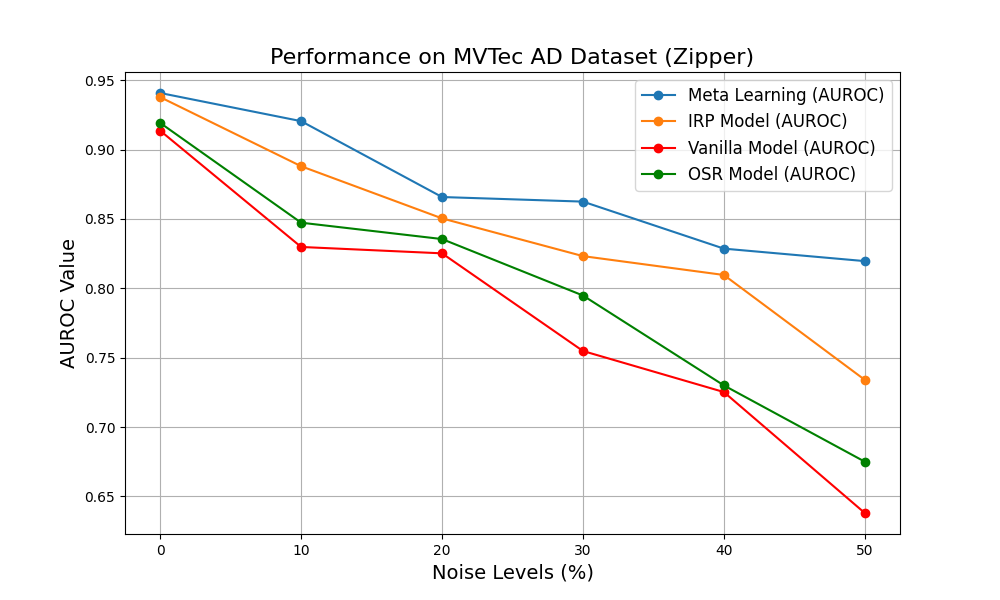}\label{fig:zipper}}
\caption{AUROC performance across various classes of the MVTec-AD dataset.}
\label{fig:mvtecclasses2}
\end{figure}

We conducted a comparative analysis of four distinct anomaly detection models: the Meta Learning Model, IPR Model, Vanilla Model, and OSR Model, across ten classes of the MVTec AD dataset: Bottle, Cable, Capsule, Carpet, Leather, MetalNut, Pill, Screw, Tile, and Zipper. The performance of these models was evaluated under different noise conditions to understand each model’s resilience to noise interference and its capability to handle diverse industrial defects. The comparative performance across these classes is demonstrated in Figure~\ref{fig:mvtecclasses2} showcasing the AUROC results for each model under various noise scenarios.

The Meta Learning Model exhibited exceptional performance across all classes, maintaining high AUROC values even in high noise environments, showcasing its robustness and reliability for industrial applications. Notably, in low noise settings, this model performed outstandingly well, highlighting its potential in precision-critical applications.
In particular, we can appreciate how our model, proposed to handle noisy data, outperforms the baseline method also in case of no noisy data, \ie noise level 0\%. We hypothesize that this effect is due to the ability of our method to effectively isolate those samples that, despite being labeled as nominal, present some aspect typical of anomalous data, pushing the boundaries of the learned model towards the anomalous region (see Figure \ref{fig:ourmethod}) and thus generating more false negatives at testing time.

In contrast, the IPR Model showed strong initial performance at low noise levels but experienced a significant drop in effectiveness as noise increased. This trend was particularly marked in classes such as Cable and Capsule. The Vanilla Model, though stable initially, demonstrated a sharper decline in AUROC with increased noise levels, particularly struggling with complex defect patterns in the MetalNut and Zipper classes. The OSR Model, initially outperforming the Vanilla Model, also showed a marked decrease in performance with rising noise levels, indicating its lesser robustness to data variability. The detailed AUROC scores for each class at various noise levels are summarized in Table~\ref{tab:performance}. This table provides a comprehensive statistical breakdown, including standard deviation metrics that offer further insights into the variability and reliability of each model under different testing conditions.

\begin{table}[ht]
\centering
\caption{Detailed AUROC scores for each model and class at varying noise levels.}
\label{tab:performance}
\resizebox{\textwidth}{!}{%
\begin{tabularx}{\textwidth}{@{}X*{6}{>{\centering\arraybackslash}X}@{}}
\toprule
& \multicolumn{6}{c}{\textbf{Noise Level (\%)}} \\
\cmidrule(lr){2-7}
\textbf{Class} & \textbf{0\%} & \textbf{10\%} & \textbf{20\%} & \textbf{30\%} & \textbf{40\%} & \textbf{50\%} \\
\midrule
\textbf{Bottle}          & 0.9493 & 0.9442 & 0.9363 & 0.9318 & 0.9393 & 0.9354 \\
\textbf{Cable}           & 0.9309 & 0.8860 & 0.8776 & 0.8409 & 0.8317 & 0.8072 \\
\textbf{Capsule}         & 0.8953 & 0.8655 & 0.8425 & 0.8397 & 0.8241 & 0.8173 \\
\textbf{Carpet}          & 0.8499 & 0.7986 & 0.7705 & 0.6886 & 0.6794 & 0.7064 \\
\textbf{Leather}         & 0.9746 & 0.9522 & 0.9419 & 0.9414 & 0.9352 & 0.9180 \\
\textbf{MetalNut}        & 0.9133 & 0.8803 & 0.8687 & 0.8512 & 0.8201 & 0.8133 \\
\textbf{Pill}            & 0.8973 & 0.8709 & 0.8481 & 0.8373 & 0.8173 & 0.8104 \\
\textbf{Screw}           & 0.9004 & 0.8747 & 0.8489 & 0.8368 & 0.8278 & 0.8259 \\
\textbf{Tile}            & 0.9848 & 0.9874 & 0.9723 & 0.9749 & 0.9622 & 0.9522 \\
\textbf{Zipper}          & 0.9408 & 0.9205 & 0.8658 & 0.8625 & 0.8286 & 0.8196 \\
\textbf{MVTec-AD}        & 0.9237 & 0.8980 & 0.8773 & 0.8605 & 0.8466 & 0.8406 \\
\textbf{KSDD2}           & 0.9428 & 0.9340 & 0.9160 & 0.9182 & 0.9108 & 0.9027 \\
\textbf{Standard Deviation}  & $\pm$0.0386 & $\pm$0.0512 & $\pm$0.0566 & $\pm$0.0750 & $\pm$0.0779 & $\pm$0.0779 \\
\bottomrule
\end{tabularx}}
\end{table}

\subsection{Overall Performance Analysis}
\begin{figure}[t]
\centering
\subfloat[MVTec AD]{\includegraphics[width=0.49\textwidth]{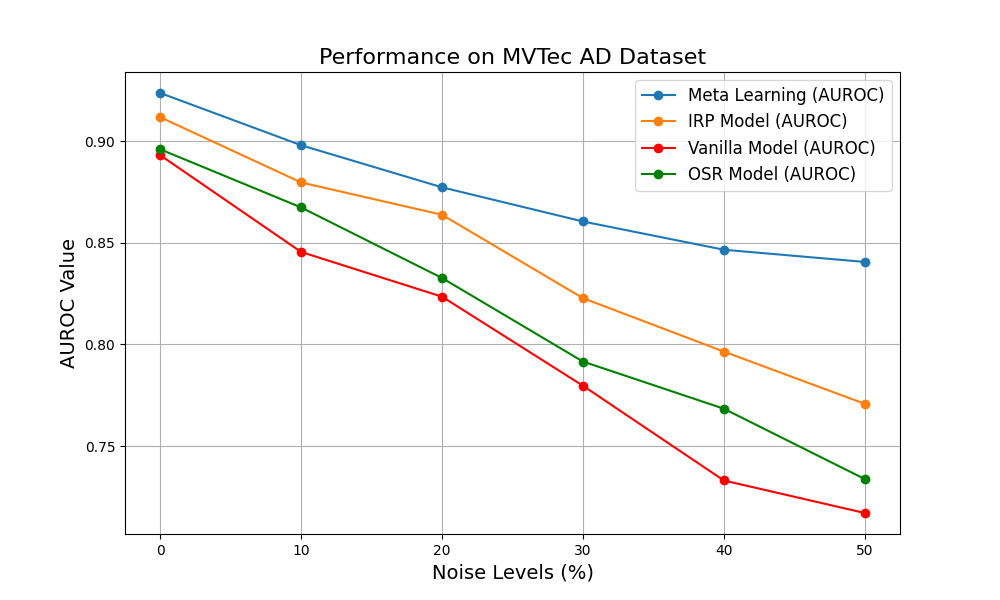}\label{fig:mvtec}}
\hfill
\subfloat[KSDD2]{\includegraphics[width=0.49\textwidth]{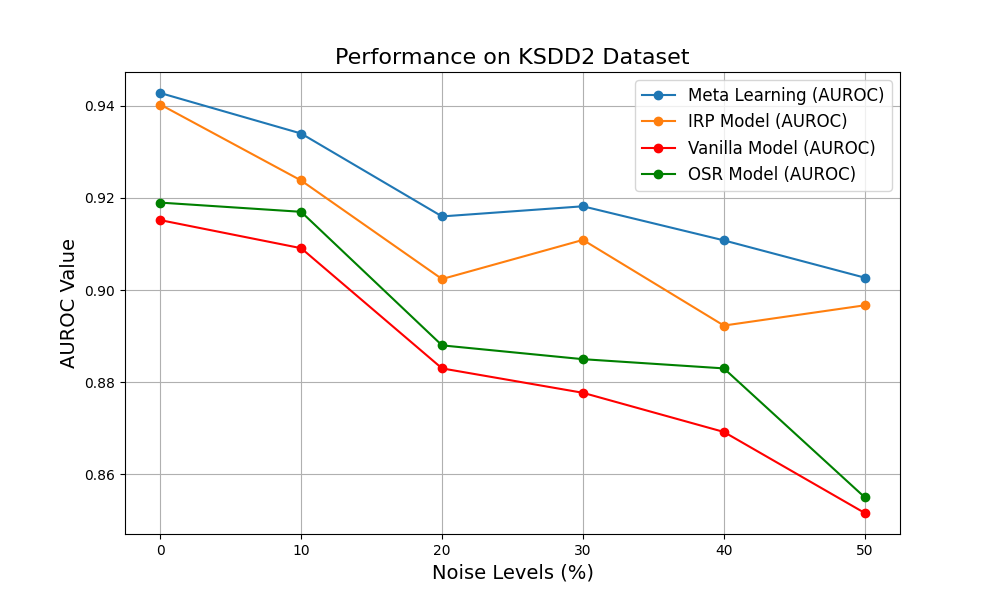}\label{fig:ksdd2}}
\caption{Overall results on MVTec-AD and KSDD2 datsets.}
\label{fig:datasets}
\end{figure}

Figure~\ref{fig:mvtec} aggregates the findings from the MVTec AD dataset, presenting the overall effectiveness of each model. The Meta Learning Model stands out with consistently high AUROC scores, reaffirming its robustness and adaptability in diverse conditions. This figure~\ref{fig:ksdd2} illustrates the performance results on the KSDD2 dataset, further demonstrating the models' effectiveness across additional industrial anomalies.

The overall results on the MVTec-AD and KSDD2 datasets, as depicted in Figure~\ref{fig:datasets}, confirm the Meta Learning Model as a superior choice for robust anomaly detection, suitable for industrial applications where both precision and adaptability are crucial. The examination of AUROC performance across different classes and noise levels highlights the importance of model selection in industrial anomaly detection. While each model exhibits unique strengths, the consistently high performance of the Meta Learning Model across various conditions and datasets recommends it as the optimal choice for practical implementations where reliability and accuracy are paramount.

This comprehensive evaluation benchmarks the capabilities of these models and provides a valuable framework for further refinement of anomaly detection technologies tailored to specific industrial requirements.

\subsection{Quantitative Analysis of Deleted Samples}

In this analysis, we explore the number of samples removed during the iterative refinement process with the KSDD2 and MVTec-AD datasets, as presented in Table~\ref{tab:performance_metrics}. The table details the count of "Good" and "Bad" images deleted at various noise levels (0\% to 50\%) across different classes such as Bottle, Cable, Capsule, Carpet, Leather, MetalNut, Pill, Screw, Tile, and Zipper. At a 0\% noise level, the data shows that only "Good" images are removed. This pattern suggests that these non-defective samples might possess characteristics that cause them to be initially misidentified as defective. These images can be associated with samples with low likelihood in a Gaussian distribution. While this is not a problem in an infinite population, in case of the limited amount of data, they can easily lead to overestimating the standard deviation of the distribution thus producing classification boundaries that are prone to generate false negatives, \ie anomalous samples not recognized as that. 

As the noise increases, more "Bad" images are deleted across all classes, demonstrating the model's effectiveness in identifying true defects under challenging conditions. It is worth noting that at 10\% noise level, over 9 "Bad" images are deleted on average, which means that only less than 10\% of the anomalous images survive the filtering process. For example, the Carpet class starts with 24 good images being removed at 0\% noise, with a progressive increase in defective image removal as noise rises, reaching 30 by 50\% noise. This trend is consistent across other categories, indicating robust model performance in various noise environments. This data provides insights into the model's capabilities to distinguish between defective and non-defective samples effectively, ensuring reliability and accuracy in practical applications.

\begin{table}[ht]
    \centering
    \caption{Number of images deleted at various noise levels for KSDD2 and MVTec-AD datasets, segmented by product categories. The table shows the counts of deleted images categorized as "Good" and "Bad" for each noise level across different classes.}
    \label{tab:performance_metrics}
    \resizebox{\textwidth}{!}{%
    \begin{tabular}{@{}lcccccccccccccccccccc@{}}
    \toprule
    \multirow{3}{*}{\textbf{Class}} & \multicolumn{2}{c}{\textbf{0\%}} & \multicolumn{2}{c}{\textbf{10\%}} & \multicolumn{2}{c}{\textbf{20\%}} & \multicolumn{2}{c}{\textbf{30\%}} & \multicolumn{2}{c}{\textbf{40\%}} & \multicolumn{2}{c}{\textbf{50\%}} \\
    \cmidrule(lr){2-3} \cmidrule(lr){4-5} \cmidrule(lr){6-7} \cmidrule(lr){8-9} \cmidrule(lr){10-11} \cmidrule(lr){12-13}
    & \textbf{Good} & \textbf{Bad} & \textbf{Good} & \textbf{Bad} & \textbf{Good} & \textbf{Bad} & \textbf{Good} & \textbf{Bad} & \textbf{Good} & \textbf{Bad} & \textbf{Good} & \textbf{Bad} \\
    \midrule
    \textbf{Bottle}   & 23 & 0 & 21 & 10 & 9  & 18 & 9  & 21 & 6  & 25 & 15 & 26 \\
    \textbf{Cable}    & 9  & 0 & 6  & 9  & 10 & 11 & 4  & 12 & 3  & 18 & 4  & 12 \\
    \textbf{Capsule}  & 14 & 0 & 10 & 8  & 11 & 13 & 8  & 19 & 8  & 23 & 7  & 25 \\
    \textbf{Carpet}   & 24 & 0 & 28 & 10 & 25 & 19 & 19 & 29 & 17 & 37 & 12 & 30 \\
    \textbf{Leather}  & 15 & 0 & 18 & 9  & 21 & 14 & 9  & 19 & 12 & 13 & 8  & 19 \\
    \textbf{MetalNut} & 18 & 0 & 16 & 8  & 11 & 15 & 12 & 17 & 4  & 24 & 6  & 18 \\
    \textbf{Pill}     & 22 & 0 & 13 & 9  & 7  & 19 & 8  & 25 & 5  & 28 & 12 & 28 \\
    \textbf{Screw}    & 20 & 0 & 13 & 10 & 10 & 9  & 7  & 22 & 6  & 21 & 8  & 32 \\
    \textbf{Tile}     & 19 & 0 & 16 & 9  & 9  & 17 & 9  & 10 & 7  & 14 & 10 & 13 \\
    \textbf{Zipper}   & 21 & 0 & 17 & 10 & 14 & 14 & 9  & 22 & 8  & 20 & 5  & 19 \\
    \textbf{KSDD2}    & 25 & 0 & 22 & 10 & 13 & 16 & 12 & 18 & 13 & 21 & 16 & 26 \\
    
    \bottomrule
    \end{tabular}}
\end{table}

\section{Conclusion}
\label{sec:conclusion}
In this paper, we presented a novel framework for anomaly detection that, leveraging the adaptation power of meta-learning and outlier rejection of statistical tools, can improve the performance of existing unsupervised anomaly detection methods by pruning out anomalous samples at training time and then fitting a model on the subset of remaining samples. 
Experiments on two popular image anomaly detection datasets demonstrate the substantial benefits of advanced meta-learning techniques for anomaly detection in industrial settings, outperforming state-of-the-art approaches across various noise levels and industrial contexts, as evidenced by its consistently high AUROC scores. This model not only proved robust in high-noise environments but also excelled in low-noise settings, highlighting its potential for precision-critical applications. 

Moving forward, further research could explore the integration of additional adaptive algorithms to improve the granularity of anomaly detection. Moreover, expanding the dataset diversity and including real-time deployment scenarios may provide deeper insights into the operational effectiveness of these models, ensuring they meet the evolving demands of industrial applications.



\section*{Acknowledgements}
This study was carried out within the PNRR research activities of the consortium iNEST (Interconnected North-Est Innovation Ecosystem) funded by the European Union Next-GenerationEU (Piano Nazionale di Ripresa e Resilienza (PNRR) – Missione 4 Componente 2, Investimento 1.5 – D.D. 1058  23/06/2022, ECS\_00000043).

%
%
\bibliographystyle{splncs04}
\bibliography{main}
\end{document}